\documentclass[letterpaper, 10 pt, conference]{ieeeconf}  

\IEEEoverridecommandlockouts                              

\usepackage{url,makecell,graphicx,subcaption,booktabs,amssymb,amsmath,amsfonts,latexsym,graphicx,float,adjustbox,multirow}

\title{KAN-RCBEVDepth: Integrating Multi-modal Sensor Data for Robust 3D Object Detection in Autonomous Driving}

\author{Zhihao Lai$^{1}$, Chuanhao Liu$^{2}$, Shihui Sheng$^{3}$, and Zhiqiang Zhang$^{4}$
\thanks{*This work was not supported by any organization}
\thanks{$^{1}$Zhihao Lai is with the School of Robotics, Xi’an Jiaotong-Liverpool University, Suzhou, Jiangsu 215123, PR China
        {\tt\small zhihao.lai23@student.xjtlu.edu.cn}}%
\thanks{$^{2}$Chuanhao Liu is with the International Business School Suzhou (IBSS), Xi’an Jiaotong-Liverpool University, Suzhou, Jiangsu 215123, PR China
        {\tt\small chuanhao.liu23@student.xjtlu.edu.cn}}%
\thanks{$^{3}$Shihui Sheng is with the International Business School Suzhou (IBSS), Xi’an Jiaotong-Liverpool University, Suzhou, Jiangsu 215123, PR China
        {\tt\small shihui.sheng23@student.xjtlu.edu.cn}}%
\thanks{$^{4}$Zhiqiang Zhang is with the College of Information and Communication Engineering, Communication University of China, Beijing 100024, PR China
        {\tt\small zzq\_cuc@cuc.edu.cn}}%
}

\begin{document}

\maketitle

\begin{abstract}
Accurate 3D object detection in autonomous driving is critical yet challenging due to occlusions, varying object sizes, and complex urban environments. This paper introduces the KAN-RCBEVDepth method, an innovative approach aimed at enhancing 3D object detection by fusing multimodal sensor data from cameras, LiDAR, and millimeter-wave radar. Our unique Bird's Eye View-based approach significantly improves detection accuracy and efficiency by seamlessly integrating diverse sensor inputs, refining spatial relationship understanding, and optimizing computational procedures. Experimental results show that the proposed method outperforms existing techniques across multiple detection metrics, achieving a higher Mean Distance AP (0.389, 23\% improvement), a better ND Score (0.485, 17.1\% improvement), and a faster Evaluation Time (71.28s, 8\% faster). Additionally, the KAN-RCBEVDepth method significantly reduces errors compared to BEVDepth, with lower Transformation Error (0.6044, 13.8\% improvement), Scale Error (0.2780, 2.6\% improvement), Orientation Error (0.5830, 7.6\% improvement), Velocity Error (0.4244, 28.3\% improvement), and Attribute Error (0.2129, 3.2\% improvement). These findings suggest that our method offers enhanced accuracy, reliability, and efficiency, making it well-suited for dynamic and demanding autonomous driving scenarios.  The code will be released in \url{https://github.com/laitiamo/RCBEVDepth-KAN}.
\end{abstract}


\section{Introduction}

Accurate 3D object detection is a critical component of autonomous driving systems, enabling vehicles to perceive their environment in three dimensions and precisely identify and localize surrounding objects such as vehicles, including vehicles, pedestrians, and obstacles\cite{liu2020multi}. In urban environments, where visibility is often limited and object density is high, accurate 3D object detection is essential for recognizing occluded objects and improving object localization, thereby enhancing situational awareness and facilitating safer navigation.

\begin{figure}[t]
    \centering
    \includegraphics[width=\linewidth]{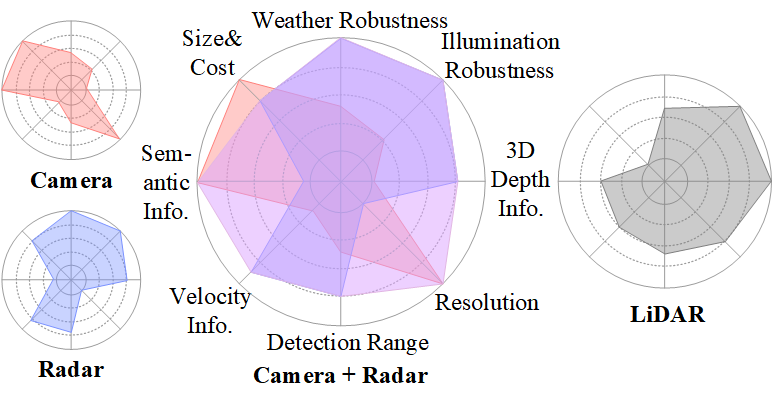}
    \caption{Illustration of the complementary nature of camera and radar sensors}
    \label{fig:camera-radar-complement}
\end{figure}

To address the limitations of individual sensors, integrating data from cameras, light detection and ranging radar (LiDAR), and radar via sensor fusion can provide a more comprehensive and accurate perception of the environment. As illustrated in Figure \ref{fig:camera-radar-complement}, these sensors' complementary strengths—visual detail, weather resilience, and precise depth data—work together to handle diverse environmental conditions effectively\cite{Singh2023}. Cameras offer detailed visuals and precise object boundaries, while millimeter-wave radar performs well in adverse weather conditions and efficiently estimates object depth and velocity using the Doppler effect. LiDAR delivers high-precision 3D point cloud data crucial for accurate depth perception.  By leveraging the strengths of each sensor type, sensor fusion mitigates their weaknesses, thereby enhancing the overall performance of 3D object detection systems.

Traditional multimodal fusion at the perception result level often limits performance. However, integrating modalities into a unified feature space, such as with Bird's Eye View (BEV) perception, can overcome these challenges. Current research focuses on efficiently processing sparse, unordered point cloud data. LiDAR-based 3D object detection techniques are primarily categorized into voxel and pillar methods. VoxelNet\cite{zhou2018voxelnet} organizes unordered point cloud data into structured voxels and uses 3D convolutional networks to extract features. CenterFusion\cite{nabati2021} detects objects without relying on fixed-size anchor boxes by assigning center points. Voxel\cite{deng2021voxel} R-CNN enhances point cloud features through voxel pooling.

\begin{figure*}[t]
    \centering
    \includegraphics[width=0.9\linewidth]{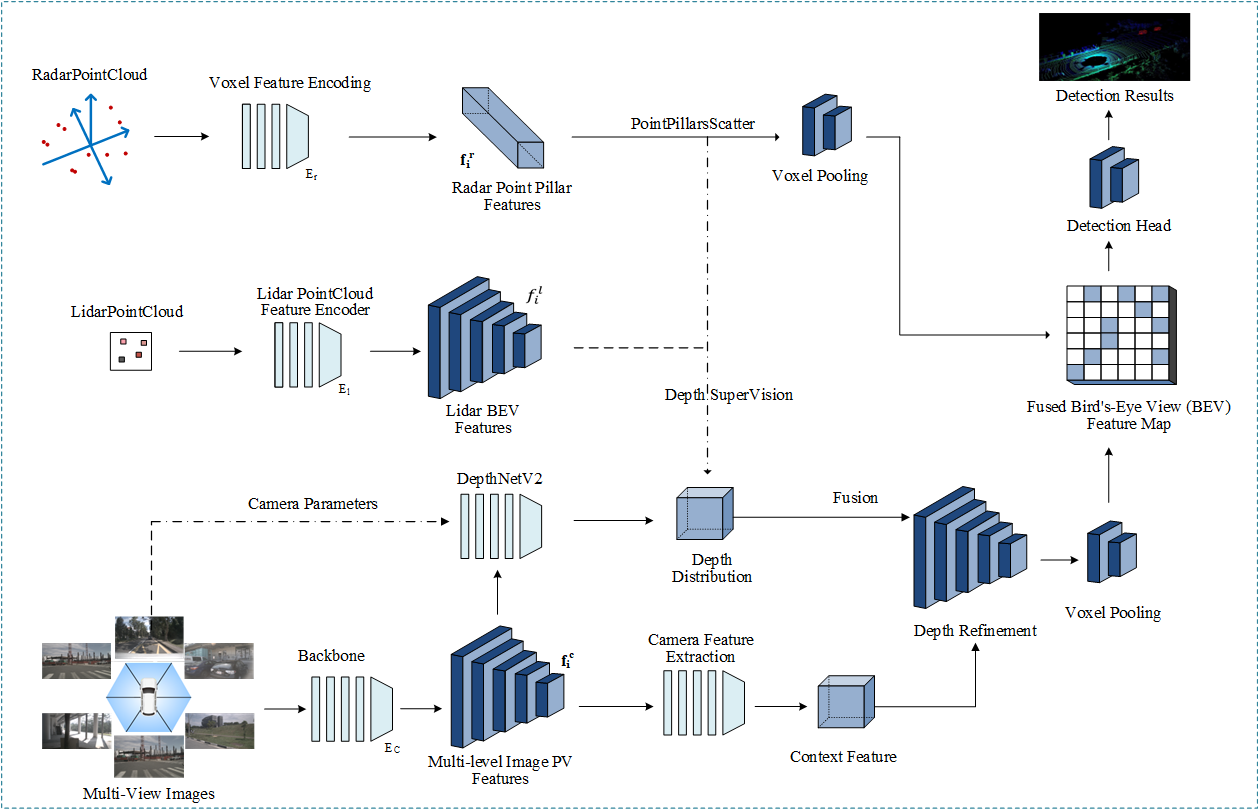} 
    \caption{System architecture of the KAN-RCBEVDepth framework highlighting the integration of multimodal sensors and depth supervision for accurate depth estimation.} 
    \label{fig:KAN-RCBEVDepth-architecture} 
\end{figure*}

As noted in BEVDepth\cite{li2022bevdepth}, the Lift-Splat method  encounters challenges due to inaccurate depth estimation, which results in only a subset of features being correctly re-projected into the BEV space, leading to imprecise semantic information. Implementing explicit depth supervision can mitigate this issue, thereby enhancing the overall performance of 3D object detection. 

In this study, we introduce KAN-RCBEVDepth, a novel framework for 3D object detection in autonomous vehicles that integrates data from LiDAR, millimeter-wave radar, and cameras into a cohesive BEV space. This integration optimizes feature fusion through efficient voxel pooling, enhanced by the DepthNetV2 which employs a tokenized Kolmogorov-Arnold Network(KAN) for precise depth estimation and robust feature processing. As illustrated in Figure \ref{fig:KAN-RCBEVDepth-architecture}, this framework uses LiDAR and radar for explicit depth supervision. Our contributions are summarized as follows:

\begin{itemize}
    \item We designed the Multi-View Feature Fusion Unit (MFFU) to integrate camera images and millimeter-wave radar point clouds into a shared BEV space, creating a unified representation where features from different views coexist seamlessly, allowing for effective feature fusion and task-specific outputs.
    \item We extended a supervision module to utilize ground-truth depth data from both LiDAR and millimeter-wave radar point clouds. In scenarios where LiDAR data is sparse, radar data provides crucial supplementary information. A depth refinement module further improves depth accuracy by fine-tuning unprojected features. 
    \item Camera features are transformed using intrinsics and extrinsics. We use a tokenized KAN block and a Squeeze-and-Excitation module to enhance DepthNet, improving feature extraction and representation. 
    \item We employ CUDA-accelerated voxel pooling to efficiently voxelize radar point clouds, integrating these into the BEV feature map. DepthNetV2 concurrently processes image features, which are then fused with voxelized radar data for a comprehensive BEV representation.
\end{itemize}

\section{Related Works}

\begin{figure*}[ht]
    \centering
    \includegraphics[width=1\linewidth]{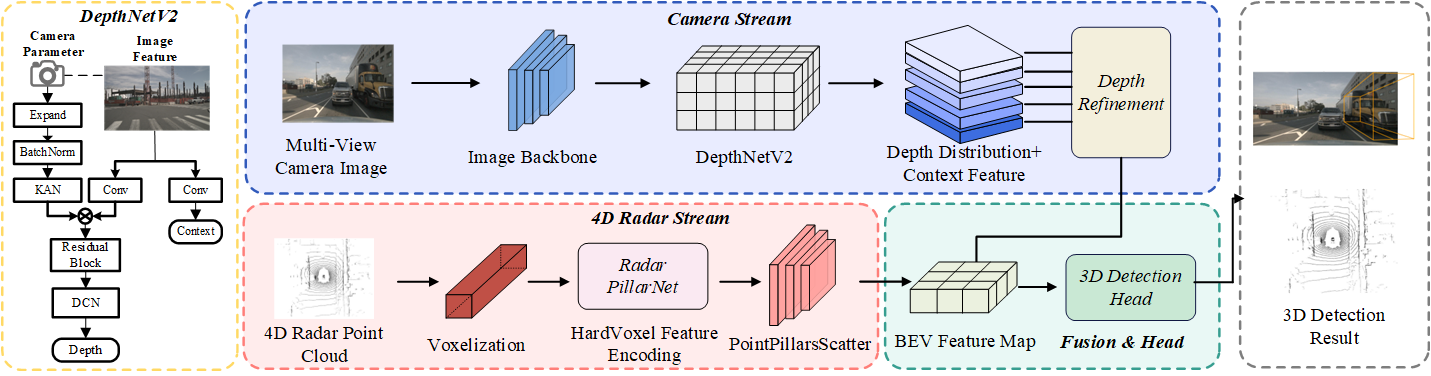}
    \caption{MFFU integrated with the improved depth network. The left half of the figure illustrates the framework of the depth network, where context features are generated directly from image features, and a variant of the SE-like layer is designed to aggregate with image features for enhanced depth estimation. The right half showcases the MFFU for integrating radar and camera data.}
    \label{fig:MFFU}
\end{figure*}
\textbf{Multimodal Sensor Fusion-based 3D Object Detection Methods: }Early studies separately processed LiDAR and camera data, then fused them at the decision level, as seen in BEVDet\cite{huang2021bevdet}. BEV-based 3D perception, known for rich semantic and precise location information, has become vital for tasks like behavior prediction and motion planning. From 2020 to 2022, research shifted towards converting perspective view to BEV for image-based 3D object detection. Zhang et al. introduced BEVerse, which extracts spatiotemporal BEV features from multi-view cameras for multi-task inference\cite{zhang2022beverse}. Tesla used surround cameras for BEV object detection, enhancing visual 3D object detection accuracy\cite{tesla2022aiday}.

Association Modality Fusion focuses on finding spatial relationships among multi-modal sensors, primarily combining LiDAR and cameras. Approaches like Pointpainting\cite{Chen2017}, which maps image semantic labels onto LiDAR point clouds, enhance point clouds for use with any 3D detector. Similarly, MVDNet\cite{Chadwick2019} fuses radar and LiDAR data, beneficial in foggy conditions. Despite the computational cost of 2D convolution on radar points, MV3D\cite{Qian2021} integrates LiDAR into RGB channels for 3D region creation. Methods like AVOD\cite{bansal2020pointillism} enhance detection by mapping 2D boxes into 3D, while BEVFusion\cite{Liu2022} innovates by integrating depth probabilities from image features into BEV for pseudo-3D projections. Additionally, feature-level fusion methods have been proposed, such as the deep fusion method for LiDAR and vision sensors introduced by Li et al.\cite{li2022voxel}, and SupFusion\cite{qin2023supfusion}, a supervised LiDAR-camera fusion technique that enhances detection accuracy and understanding of the 3D environment.

Recent multimodal sensor fusion approaches have taken another direction. Yin et al. from the University of Texas proposed the MVP\cite{li2022deepfusion} network, which uses semantic segmentation from 2D images to generate virtual points, enhancing point cloud density. Similarly, Vora et al. from nuTonomy developed the PointPainting network, enhancing LiDAR point clouds for improved 3D detection\cite{decourt2022recurrent}.

\section{Methodology}

This section introduces key components of the KAN-RCBEVDepth framework: the MFFU for radar and camera data integration, the Improved Camera-awareness Module for depth estimation using LiDAR and radar, the Voxel Pooling Optimization for efficient BEV feature aggregation, and the Detection Head Feature Fusion for generating accurate 2D bounding boxes and improving object detection. These components collectively enhance the system's perception, efficiency, and accuracy in dynamic environments.

\subsection{Multimodal Feature Fusion Unit for Radar and Camera Data Integration}
To further enhance the system's perception capabilities, we designed the MFFU in Figure \ref{fig:MFFU}, which integrates data from multi-view cameras and millimeter-wave radar into a unified BEV feature space, including radar point cloud voxelization, HardVoxel Feature Encoding, and feature projection and extraction via PointPillarsScatter.

\subsubsection{4D Radar Stream}
In the feature encoding stage, we expand the original 4D radar point cloud data into 9 dimensions to better represent the features within each voxel. Initialy, the point cloud data is voxelized, dividing it into a grid of voxels, with each voxel containing a fixed number of points. This process converts continuous point cloud data into discrete voxel representations, simplifying subsequent feature processing.

For each voxel, we calculate the mean position of all points (cluster center) and the geometric center of the voxel. The cluster center is calculated as follows:

\begin{equation}
\text{Cluster Center} = \frac{1}{N} \sum_{i=1}^{N} P_i
\end{equation}

where \( P_i \) represents the coordinates of each point, and \( N \) is the number of points in the voxel. We also compute the distance of each point from the voxel center:

\begin{equation}
\text{Distance to Voxel Center} = \lVert P_i - C_{\text{voxel}} \rVert
\end{equation}

where \( C_{\text{voxel}} \) is the voxel's geometric center. These features, combined with the original 4D features (coordinates and reflectivity), result in a 9-dimensional feature vector for each point:

\begin{equation}
(x, y, z, r) \rightarrow (x, y, z, r, x_c, y_c, z_c, x_p, y_p)
\end{equation}

Next, space is divided in the \( x \) and \( y \) directions, creating \( H \times W \) pillars. Each pillar contains a varying number of points. To standardize processing, we set a fixed value \( T \). For each pillar, if the number of points exceeds \( T \), we randomly sample to \( T \) points; if the number of points is less than \( T \), we pad with zeros to \( T \) points. Each pillar is thus represented as a matrix of dimensions \( T \times 9 \).

\textbf{Feature Extraction:} We use the Voxel Feature Encoding (VFE) network to extract features from the points within each pillar. For each pillar, the VFE network produces a feature output of dimensions \( T \times C \), where \( C \) is the extracted feature dimension. Max pooling is then applied across \( T \) to obtain a feature vector of dimension \( C \):

\begin{equation}
\mathbf{f}_{\text{pillar}} = \max(\text{VFE}(T \times 9))
\end{equation}

Thus, each pillar results in a \( C \)-dimensional feature vector.

\textbf{Pseudo Image Construction:} All pillar feature vectors are mapped onto an \( H \times W \) space, ultimately forming a pseudo image of size \( C \times H \times W \).

\textbf{PointPillarsScatter:} After feature extraction, these features are mapped onto the corresponding pixel positions of the BEV feature map. The index for placing voxel features into the BEV map is calculated using the following formula:

\begin{equation}
\text{Index} = y \times N_x + x
\end{equation}

where \( N_x \) is the width of the BEV map, and \( x \) and \( y \) are the spatial coordinates of the voxel. This step ensures that the features are spatially aligned correctly.

We use the KAN block to replace the traditional MLP, where the B-spline branch provides tailored activations, and the shortcut branch enhances non-linearity and information flow. This allows for more adaptive responses to input variations. KAN transforms the camera parameter dimensions from \([6, 27]\) to \([6, 512, 1, 1]\), which are then used as attention weights in a SE layer applied directly to the image features. A \(1 \times 1\) convolution follows for channel scaling, producing Context Features and Depth Features.

The operation of the KAN block in transforming and fusing camera parameters with image features is formalized as:
\begin{equation}
D_i^{\text{pred}} = \psi(\text{SE}(F_i^{2d} | \text{KAN}(\xi(R_i) \oplus \xi(t_i) \oplus \xi(K_i))))
\end{equation}
where \(D_i^{\text{pred}}\) represents the predicted depth map for the \(i\)th image, \(F_i^{2d}\) is the 2D image feature extracted from the ResNet backbone, \(\xi(\cdot)\) denotes the feature embedding function applied to the camera's intrinsic and extrinsic parameters, \(\oplus\) represents the concatenation operation, and \(\psi\) is the final nonlinear transformation function, such as a convolutional layer.

The Depth Features undergo a softmax operation to generate the Depth Distribution, representing the probability distribution of depth values across the image:
\begin{equation}
P_d = \text{softmax}(D), \quad D \in \mathbb{R}^{C_D \times H \times W}
\end{equation}
where \(P_d \in \mathbb{R}^{C_D \times H \times W}\) represents the probability distribution of depth values.

Next, the Context Features, denoted as \(C \in \mathbb{R}^{C_C \times H \times W}\), are combined with the Depth Distribution \(P_d\) using an outer product:
\begin{equation}
F_{out}(i, j, k, l) = C(i, j, k) \cdot P_d(l, j, k)
\end{equation}
where \(F_{out}\) has dimensions \(C_C \times C_D \times H \times W\), effectively integrating both semantic and depth information.

To further enhance depth estimation accuracy, explicit depth supervision involves projecting both LiDAR and radar point clouds into the camera coordinate system:
\begin{equation}
\hat{P}_i^{img}(u_d, v_d, d) = K_i (R_i P + t_i).
\end{equation}
A Deformable Convolutional Network (DCN) further improves depth estimation. The aligned LiDAR and radar depth information is used to generate a more comprehensive supervised depth map:
\begin{equation}
D_{gt} = \phi(P_i^{img}),
\end{equation}
which serves as the ground truth for supervising the predicted depth \(D_{pred}\).

By integrating intrinsic and extrinsic camera parameters, the module achieves a precise depth map and enriched context features. Radar pillar features, processed through a Voxel Feature Encoding network, are represented as \(f_{r}\). The combined features are then fed into the Depth Refinement Module, which reshapes the feature map \(F_{3d}\) from \([C_F, C_D, H, W]\) to \([C_F \times H, C_D, W]\) and applies 3×3 convolutions along the depth axis. This module employs Voxel Pooling to aggregate features across depth planes, broadening the receptive field and significantly improving depth prediction accuracy.

\subsection{Voxel Pooling Optimization Process}

\begin{figure}[t]
    \centering
    \includegraphics[width=0.9\linewidth]{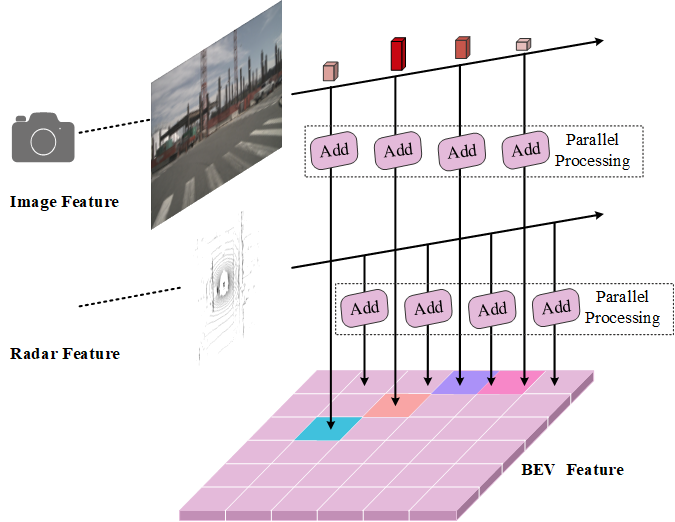}
    \caption{Details of Voxel Pooling Optimization Process}
    \label{fig:voxel}
\end{figure}

In Figure \ref{fig:voxel}, the features are filled into the corresponding positions on the BEV feature map. Millimeter-wave radar and camera data are first voxelized to generate the corresponding features. These features are then positioned on the BEV feature map, integrating data from different sensors into a unified representation.
\begin{table*}[t]
    \centering
    \begin{minipage}{0.47\textwidth}
        \centering
        \caption{\small Comparison of Detection Results and Mean Distance AP for BEVDepth and KAN-RCBEVDepth}
        \label{tab:detection_results}
        \renewcommand{\arraystretch}{1.1} 
        \resizebox{\textwidth}{!}{
            \begin{tabular}{@{}c@{\hskip 5pt}|@{\hskip 5pt}c@{\hskip 5pt}|@{\hskip 5pt}c@{\hskip 5pt}|@{\hskip 5pt}c@{\hskip 5pt}|@{\hskip 5pt}c@{\hskip 5pt}|@{\hskip 5pt}c@{\hskip 5pt}|@{\hskip 5pt}c@{\hskip 5pt}|@{\hskip 5pt}c@{\hskip 5pt}|@{\hskip 5pt}c@{}}
                \hline
                \textbf{Object} & \textbf{Method} & \textbf{0.5m} & \textbf{1.0m} & \textbf{2.0m} & \textbf{4.0m} & \textbf{mAP} \\
                \hline
                \small Car & \small BEVDepth & \small 0.152 & \small 0.405 & \small 0.641 & \small 0.734 & \small 0.483 \\
                & \small KAN-RCBEVDepth & \small 0.320 & \small 0.622 & \small 0.761 & \small 0.809 & \small 0.628 \\
                \hline
                \small Truck & \small BEVDepth & \small 0.015 & \small 0.129 & \small 0.352 & \small 0.509 & \small 0.252 \\
                & \small KAN-RCBEVDepth & \small 0.059 & \small 0.239 & \small 0.462 & \small 0.575 & \small 0.334 \\
                \hline
                \small Bus & \small BEVDepth & \small 0.024 & \small 0.224 & \small 0.509 & \small 0.684 & \small 0.360 \\
                & \small KAN-RCBEVDepth & \small 0.098 & \small 0.373 & \small 0.659 & \small 0.730 & \small 0.465 \\
                \hline
                \small Trailer & \small BEVDepth & \small NaN & \small 0.033 & \small 0.198 & \small 0.386 & \small 0.154 \\
                & \small KAN-RCBEVDepth & \small NaN & \small 0.063 & \small 0.316 & \small 0.444 & \small 0.206 \\
                \hline
                \footnotesize \textbf{Construction} \textbf{Vehicle} & \small BEVDepth & \small NaN & \small 0.005 & \small 0.106 & \small 0.187 & \small 0.074 \\
                & \small KAN-RCBEVDepth & \small NaN & \small 0.036 & \small 0.197 & \small 0.289 & \small 0.131 \\
                \hline
                \small Pedestrian & \small BEVDepth & \small 0.113 & \small 0.236 & \small 0.331 & \small 0.394 & \small 0.268 \\
                & \small KAN-RCBEVDepth & \small 0.167 & \small 0.309 & \small 0.402 & \small 0.462 & \small 0.335 \\
                \hline
                \small Motorcycle & \small BEVDepth & \small 0.069 & \small 0.281 & \small 0.422 & \small 0.509 & \small 0.320 \\
                & \small KAN-RCBEVDepth & \small 0.148 & \small 0.411 & \small 0.518 & \small 0.549 & \small 0.406 \\
                \hline
                \small Bicycle & \small BEVDepth & \small 0.106 & \small 0.278 & \small 0.400 & \small 0.437 & \small 0.305 \\
                & \small KAN-RCBEVDepth & \small 0.170 & \small 0.370 & \small 0.449 & \small 0.477 & \small 0.366 \\
                \hline
                \small Traffic Cone & \small BEVDepth & \small 0.252 & \small 0.413 & \small 0.507 & \small 0.580 & \small 0.438 \\
                & \small KAN-RCBEVDepth & \small 0.314 & \small 0.469 & \small 0.554 & \small 0.621 & \small 0.490 \\
                \hline
                \small Barrier & \small BEVDepth & \small 0.201 & \small 0.504 & \small 0.628 & \small 0.687 & \small 0.505 \\
                & \small KAN-RCBEVDepth & \small 0.239 & \small 0.544 & \small 0.646 & \small 0.693 & \small 0.531 \\
                \hline
            \end{tabular}
        }
    \end{minipage}%
    \hfill
    \begin{minipage}{0.48\textwidth}
        \centering
        \caption{\small Comparison of Errors for BEVDepth and KAN-RCBEVDepth}
        \label{tab:error_results}
        \renewcommand{\arraystretch}{1.3} 
        \resizebox{\textwidth}{!}{ 
            \begin{tabular}{@{}c@{\hskip 5pt}|@{\hskip 5pt}c@{\hskip 5pt}|@{\hskip 5pt}c@{\hskip 5pt}|@{\hskip 5pt}c@{\hskip 5pt}|@{\hskip 5pt}c@{\hskip 5pt}|@{\hskip 5pt}c@{\hskip 5pt}|@{\hskip 5pt}c@{\hskip 5pt}|@{\hskip 5pt}c@{}}
                \hline
                \textbf{Object} & \textbf{Method} & \textbf{Trans Err} & \textbf{Scale Err} & \textbf{Orient Err} & \textbf{Vel Err} & \textbf{Attr Err} \\  
                \hline
                \small Car & \small BEVDepth & \small 0.553 & \small 0.171 & \small 0.247 & \small 0.631 & \small 0.233 \\
                & \small KAN-RCBEVDepth & \small 0.393 & \small 0.169 & \small 0.176 & \small 0.438 & \small 0.215 \\
                \hline
                \small Truck & \small BEVDepth & \small 0.751 & \small 0.227 & \small 0.291 & \small 0.580 & \small 0.227 \\
                & \small KAN-RCBEVDepth & \small 0.630 & \small 0.220 & \small 0.193 & \small 0.371 & \small 0.215 \\
                \hline
                \small Bus & \small BEVDepth & \small 0.734 & \small 0.226 & \small 0.218 & \small 1.224 & \small 0.263 \\
                & \small KAN-RCBEVDepth & \small 0.602 & \small 0.202 & \small 0.133 & \small 0.652 & \small 0.221 \\
                \hline
                \small Trailer & \small BEVDepth & \small 0.967 & \small 0.234 & \small 0.621 & \small 0.545 & \small 0.166 \\
                & \small KAN-RCBEVDepth & \small 0.903 & \small 0.242 & \small 0.588 & \small 0.275 & \small 0.158 \\
                \hline
                \footnotesize \textbf{Construction} \textbf{Vehicle} & \small BEVDepth & \small 0.999 & \small 0.509 & \small 1.251 & \small 0.123 & \small 0.361 \\
                & \small KAN-RCBEVDepth & \small 0.955 & \small 0.507 & \small 1.262 & \small 0.122 & \small 0.407 \\
                \hline
                \small Pedestrian & \small BEVDepth & \small 0.762 & \small 0.302 & \small 1.015 & \small 0.599 & \small 0.305 \\
                & \small KAN-RCBEVDepth & \small 0.652 & \small 0.293 & \small 0.901 & \small 0.586 & \small 0.263 \\
                \hline
                \small Motorcycle & \small BEVDepth & \small 0.640 & \small 0.273 & \small 0.866 & \small 0.747 & \small 0.197 \\
                & \small KAN-RCBEVDepth & \small 0.522 & \small 0.253 & \small 0.865 & \small 0.718 & \small 0.212 \\
                \hline
                \small Bicycle & \small BEVDepth & \small 0.558 & \small 0.272 & \small 0.934 & \small 0.286 & \small 0.007 \\
                & \small KAN-RCBEVDepth & \small 0.447 & \small 0.267 & \small 0.944 & \small 0.234 & \small 0.011 \\
                \hline
                \small Traffic Cone & \small BEVDepth & \small 0.535 & \small 0.353 & \small NaN & \small NaN & \small NaN \\
                & \small KAN-RCBEVDepth & \small 0.475 & \small 0.348 & \small NaN & \small NaN & \small NaN \\
                \hline
                \small Barrier & \small BEVDepth & \small 0.514 & \small 0.288 & \small 0.237 & \small NaN & \small NaN \\
                & \small KAN-RCBEVDepth & \small 0.465 & \small 0.280 & \small 0.184 & \small NaN & \small NaN \\
                \hline
            \end{tabular}
        }
    \end{minipage}
\end{table*}
During feature aggregation, sorting a large number of BEV grids introduces significant computational overhead. The traditional prefix sum calculation is sequential and inefficient, so we introduced the voxel pooling technique. The observation space (ego space) is divided into grids, each representing a voxel in the BEV feature map. CUDA threads parallelize the processing of these grids, accumulating 3D features within each voxel. Each CUDA thread processes a frustum point and updates the feature map using atomic addition, ensuring accurate updates. We employed a prefix sum trick (cumsum trick) to sort and accumulate features, then used subtraction to quickly generate accurate aggregated features on the BEV feature map. By aligning frustum features from different time frames into the current coordinate system, we eliminate the effects of self-motion on detection results. Voxel pooling is performed on these aligned features, and the fused BEV features are then input into subsequent detection tasks.

\subsection{Detection Head Feature Fusion}
The key to the feature fusion process is processing features extracted from the millimeter-wave radar (\(f_r\)) and the camera (\(f_c\)) with the Depth Refinement module. After voxel pooling, the BEV feature map (\(f_{bev}\)) is generated. The detection head, as the final processing module in the system, receives the fused BEV feature map as input, filtering the millimeter-wave radar and camera point cloud data using location prior information (\(r_i, c_i\)) obtained from heatmaps generated by the main 3D detection head; the processed point cloud then generates 2D bounding boxes in the BEV perspective, which are matched with valid areas from the heatmaps using Intersection over Union (IOU) matching. The matched positional information is used to generate tensors related to radar positional coordinates and velocity information (\(q_i\)), which are then fused with the original feature map for subsequent predictions.

The final image-based feature \( f_{\text{img}} \) is obtained by aggregating multi-scale image features using a weighted summation:
\begin{equation}
f_{\text{img}} = \sum_{j} w_j \cdot f_{img}(r_j),
\end{equation}
where \( w_j \) is the weight assigned to each image feature corresponding to the reference point \( r_j \) on the image plane.

Next, during the feature fusion process, features from the BEV map \( f_{bev} \) and radar features \( f_r \) are combined. Each point in the BEV feature map is projected to 3D reference points, and the radar features corresponding to these points are aggregated. The combined features result from a summation operation, where the final fused feature \( f_{\text{final}} \) is calculated as:
\begin{equation}
f_{\text{final}} = \sum_{i} \left( f_{bev}(p_i) + f_r(p_i) \right) + f_{\text{depth}}(p_i),
\end{equation}
where \( p_i \) denotes the projected points in the BEV map, and \( f_{\text{depth}}(p_i) \) represents the depth-related features corresponding to these points.

Finally, Binary Cross Entropy handles depth loss correction. The detection loss \(L_{det}\) sums the heatmap-based classification loss \(L_{\text{heatmap}}\) and the regression loss \(L_{\text{bbox}}\):
\begin{equation}
L_{det} = L_{\text{heatmap}} + L_{\text{bbox}}.
\end{equation}

\section{Experiments}
In this section, we introduce our experimental setup and conduct experiments to validate our proposed components. 

\textbf{Dataset:} nuScenes\cite{lang2019pointpillars}, which consists of 1000 scenes, each lasting 20 seconds. It includes a comprehensive set of 3D box annotations for 23 object classes and 8 attributes, along with a complete 360-degree sensor suite comprising lidar, images, and radar. The dataset includes approximately 1.4 million camera images, 390,000 lidar sweeps, and 1.4 million radar sweeps.


\begin{table*}[t]
    \centering
    \begin{minipage}{0.48\textwidth}
        \centering
        \caption{\small KAN-RCBEVDepth's ablation study comparing different Camera-awareness on the nuScenes val set.}
        \label{tab:ablation_study_comparison_1}
        \resizebox{\textwidth}{!}{
            \begin{tabular}{c@{\hskip 6pt}c@{\hskip 6pt}c@{\hskip 6pt}c@{\hskip 6pt}c@{\hskip 6pt}c@{\hskip 6pt}c}
                \toprule
                \makecell{DepthNet} & \makecell{Improved\\DepthNet} &  \makecell{mAP$\uparrow$} & \makecell{mATE$\downarrow$} & \makecell{mAOE$\downarrow$} & \makecell{NDS$\uparrow$} \\
                \midrule
                Camera &  & 0.301 & 0.716 & 0.647 & 0.357 \\
                 & Camera  & 0.335 & 0.675 & 0.622 & 0.384 \\
                 &  Camera+Radar & 0.389 & 0.604 & 0.583 & 0.485 \\
                \bottomrule
            \end{tabular}
        }
    \end{minipage}
    \hfill
    \begin{minipage}{0.48\textwidth}
        \centering
        \caption{\small Ablation study comparing Camera-only and Camera+Radar on the nuScenes val set.}
        \label{tab:ablation_study_comparison_2}
        \resizebox{\textwidth}{!}{
        \begin{tabular}{c@{\hskip 6pt}c@{\hskip 6pt}c@{\hskip 6pt}c@{\hskip 6pt}c@{\hskip 6pt}c@{\hskip 6pt}c}
            \toprule
            Method & MFFU & \makecell{mAP$\uparrow$} & \makecell{mATE$\downarrow$} & \makecell{mAOE$\downarrow$} & \makecell{NDS$\uparrow$} \\
            \midrule
            BEVDepth &  & 0.316 & 0.701 & 0.631 & 0.415 \\
            Ours &  & 0.356 & 0.652 & 0.612 & 0.417 \\
            Ours & \checkmark & 0.389 & 0.604 & 0.583 & 0.485 \\
            \bottomrule
        \end{tabular}
        }
    \end{minipage}
\end{table*}
\begin{table*}[h]
    \centering
    \caption{Comprehensive Comparison of mean Average Scores for BEVDepth and KAN-RCBEVDepth}
    \small
    \begin{tabular}{@{}c@{\hskip 5pt}|@{\hskip 5pt}c@{\hskip 5pt}|@{\hskip 5pt}c@{\hskip 5pt}|@{\hskip 5pt}c@{\hskip 5pt}|@{\hskip 5pt}c@{\hskip 5pt}|@{\hskip 5pt}c@{\hskip 5pt}|@{\hskip 5pt}c@{\hskip 5pt}|@{\hskip 5pt}c@{\hskip 5pt}|@{\hskip 5pt}c@{}}
        \hline
        \small Method & \small mTE & \small mSE & \small mOE & \small mVE & \small mAE & \small Precision & \small NDS & \small Eval Time \\
        \hline
        \small BEVDepth\cite{li2022bevdepth} & \small 0.7014 & \small 0.2855 & \small 0.6310 & \small 0.5919 & \small 0.2199 & \small 0.3160 & \small 0.4150 & \small 77.15 s \\
        \hline
        \small Ours & \small 0.6044 & \small 0.2780 & \small 0.5830 & \small 0.4244 & \small 0.2129 & \small 0.3891 & \small 0.4845 & \small 71.28 s \\
        \hline
    \end{tabular}
    \label{tab:analysis}
\end{table*}
\begin{table*}[t]
    \centering
    \caption{\small Comparison of 3D Object Detection Methods}
    \label{tab:3d_object_detection_comparison}
    \small
    \begin{tabular}{l@{\hskip 2pt}c@{\hskip 2pt}c@{\hskip 2pt}c@{\hskip 2pt}c@{\hskip 2pt}c@{\hskip 2pt}c@{\hskip 2pt}c@{\hskip 2pt}c@{\hskip 2pt}c@{\hskip 2pt}c}
    \hline
    \textbf{\small Methods} & \textbf{\small Image Size} & \textbf{\small Backbone} & \textbf{\small mAP$\uparrow$} & \textbf{\small NDS$\uparrow$} & \textbf{\small mATE$\downarrow$} & \textbf{\small mASE$\downarrow$} & \textbf{\small mAOE$\downarrow$} & \textbf{\small mAVE$\downarrow$} & \textbf{\small mAAE$\downarrow$} \\
            \hline
            \small KPConvPillars & \small 256x704 & \small Pillars & \small 0.049 & \small 0.139 & \small 0.823 & \small 0.428 & \small 0.607 & \small 2.081 & \small 1.000 \\ \hline
            \small BEVDet\cite{huang2021bevdet} & \small 256×704 & \small ResNet50 & \small 0.298 & \small 0.379 & \small 0.725 & \small 0.279 & \small 0.589 & \small 0.86 & \small 0.245 \\ \hline
            \small BEVerse\cite{zhang2022beverse} & \small 256×704 & \small ResNet50 & \small 0.302 & \small 0.392 & \small 0.687 & \small 0.278 & \small 0.566 & \small 0.838 & \small 0.219 \\ \hline
            \small BEVDepth\cite{li2022bevdepth} & \small 256×704 & \small ResNet50 & \small 0.316 & \small 0.415 & \small 0.701 & \small 0.286 & \small 0.631 & \small 0.592 & \small 0.220 \\ \hline
            \small BEVDet4D\cite{huang2022bevdet4d} & \small 256×704 & \small ResNet50 & \small 0.322 & \small 0.456 & \small 0.703 & \small 0.278 & \small 0.495 & \small \textbf{0.354} & \small 0.206 \\ \hline
            \small CenterFusion\cite{nabati2021} & \small 256x704 & \small DLA34 & \small 0.329 & \small 0.454 & \small 0.631 & \small 0.279 & \small 0.512 & \small 0.614 & \small \textbf{0.115} \\ \hline
            \small Fast-BEV\cite{li2024fastbev} & \small 256×704 & \small ResNet50 & \small 0.337 & \small 0.471 & \small 0.663 & \small 0.285 & \small \textbf{0.391} & \small 0.388 & \small 0.201 \\ \hline
            \small PETRv2\cite{liu2023petrv2} & \small 256×704 & \small ResNet50 & \small 0.349 & \small 0.457 & \small 0.700 & \small 0.278 & \small 0.580 & \small 0.437 & \small 0.187 \\ \hline
            \small STS\cite{wang2022sts} & \small 256×704 & \small ResNet50 & \small 0.367 & \small 0.478 & \small 0.608 & \small 0.278 & \small 0.450 & \small 0.446 & \small 0.212 \\ \hline
            \small Ours & \small 256×704 & \small ResNet50 & \small \textbf{0.389} & \small \textbf{0.485} & \small \textbf{0.604} & \small \textbf{0.278} & \small 0.583 & \small 0.4244 & \small 0.2129 \\ \hline
    \end{tabular}
\end{table*}
\textbf{Training Methods:} Experiments were conducted on an NVIDIA GeForce RTX 4070 GPU using the nuScenes dataset, including camera, lidar, and radar data. The model used ResNet50 as the backbone for processing image features, with an input resolution of 256x704, and was trained for 24 epochs. Class-Balanced Grouping and Sampling and data augmentation were applied, but exponential moving averages were not used.
\begin{figure*}[t]
    \centering
    \includegraphics[width=0.9\linewidth]{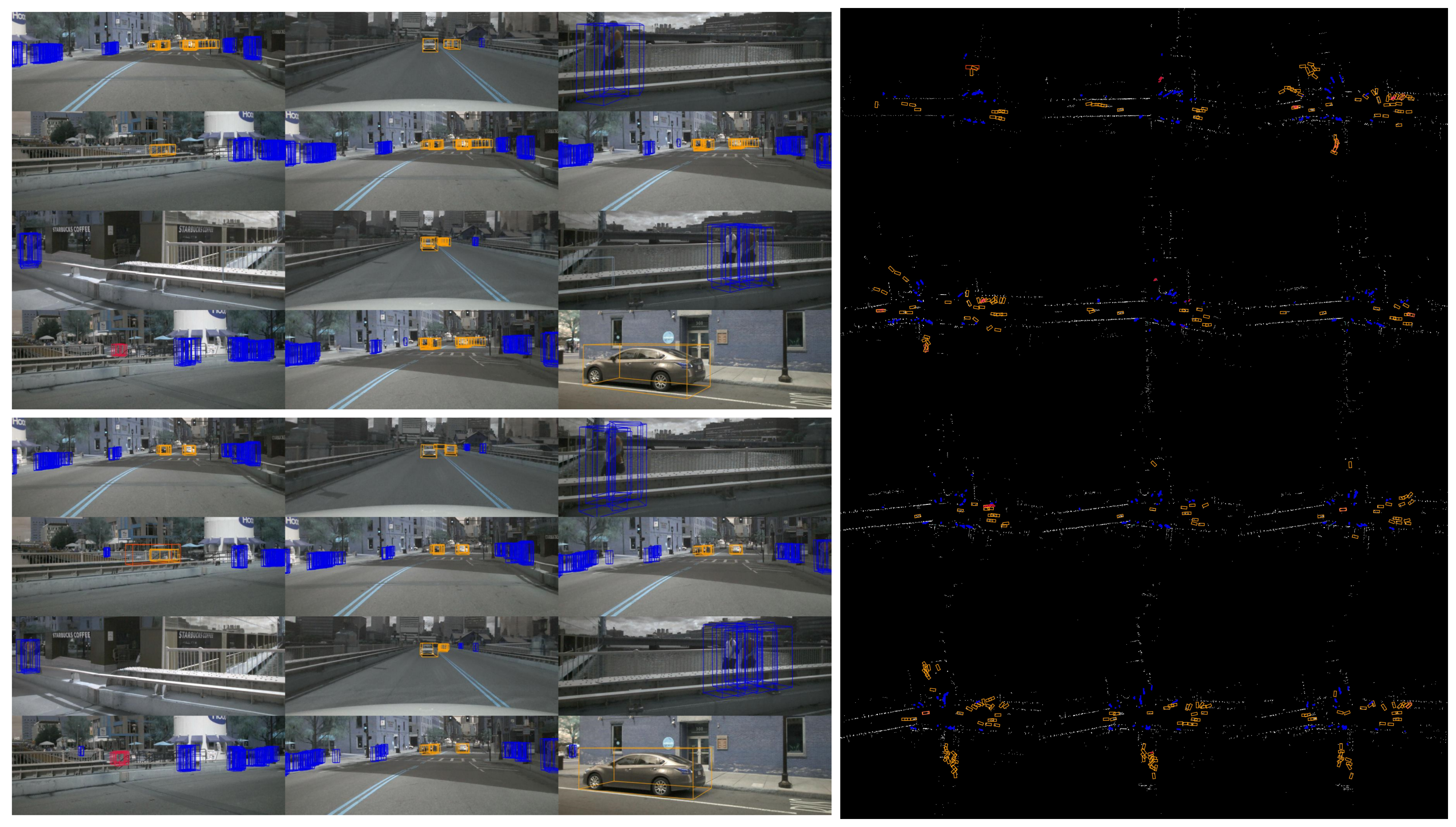} 
    \caption{Comparison of BEVDepth\cite{li2022bevdepth} and KAN-RCBEVDepth Camera and Radar Images.}
    \label{fig:camera-radar-comparison}
\end{figure*}
\subsection{Experimental Details}

By evaluating the following detection objects: Car, Pedestrian, Bicycle, Truck, Bus, Construction Vehicle, Trailer, Motorcycle, Traffic Cone, and Barrier. 

Table \ref{tab:detection_results} compares detection results and Mean Distance Average Precision (mAP) for BEVDepth\cite{li2022bevdepth} and KAN-RCBEVDepth across various objects, showing that our method outperforms BEVDepth in all categories, with mAP calculated at matching thresholds of 0.5, 1, 2, and 4 meters.

Table \ref{tab:error_results} compares key error metrics, including Translation Error (mATE), mean Scale Error (mASE), mean Orientation Error (mAOE), mean Velocity Error (mAVE), and mean Attribute Error (mAAE), where KAN-RCBEVDepth consistently shows lower errors, indicating higher accuracy in object detection and attribute estimation, particularly in these metrics.

\subsection{Ablation Experiments}

The experimental results in Table \ref{tab:ablation_study_comparison_1} demonstrate the performance comparison of awareness under different configurations, aiming to assess how adding the KAN module and incorporating multimodal sensor data impact depth estimation accuracy and object detection performance.

We compared detection performance using only the camera modality versus both camera and millimeter-wave radar modalities. The second experiment didn’t use MFFU for multimodal fusion in our method, it included millimeter-wave radar in the ground truth supervision process. Table \ref{tab:ablation_study_comparison_2} shows that adding millimeter-wave radar significantly improves mAP and NDS while reducing mATE and mAOE. These results demonstrate that incorporating the KAN block and multimodal sensor data, such as millimeter-wave radar, greatly enhances depth estimation accuracy and detection performance.

\subsection{Comparative Experiments}
Table \ref{tab:analysis} compares BEVDepth and KAN-RCBEVDepth across multiple metrics, showing that our method outperforms BEVDepth due to the MFFU's effective multimodal fusion, where KAN-RCBEVDepth achieves lower errors across all metrics. Since mAP alone cannot fully evaluate all aspects of detection tasks in the nuScenes dataset, such as velocity and attribute estimation, the ND Score (NDS) is introduced for a more comprehensive evaluation. The formula for NDS is:
\begin{equation}
NDS = \frac{1}{10} \left[ 5\text{mAP} + \sum_{\text{mTP} \in \mathcal{TP}} \left( 1 - \min(1, \text{mTP}) \right) \right]
\end{equation}

where \(mTP\) is the average true positive rate metric for a category. 

Among various algorithms compared in Table \ref{tab:3d_object_detection_comparison}, including BEVDet, BEVerse, BEVDet4D, and PETRv2, KAN-RCBEVDepth demonstrated the best performance in key metrics, achieving the highest mAP (0.3891), NDS (0.4845), and the lowest mATE (0.6044) and mASE (0.278), highlighting its superiority in detection accuracy and object position estimation.

\subsection{Qualiative Results}

Figure \ref{fig:camera-radar-comparison} compares the two main configurations. The left image shows the camera view from BEVDepth, while the top right image displays the camera view from KAN-RCBEVDepth. The bottom right image presents radar data used in KAN-RCBEVDepth. These visual comparisons highlight the enhanced detection capabilities and additional sensor data in our method, contributing to its superior performance in object detection.

\section{Conclusion}
This study explores the integration of multimodal perception technologies in autonomous driving systems, with a particular focus on advancing 3D object detection techniques for intelligent vehicles. By combining camera imagery, LiDAR, and radar data, we have developed an innovative approach using a multimodal BEV framework. Our KAN-RCBEVDepth method is particularly effective in processing multimodal data, significantly enhancing detection accuracy and robustness, even in scenarios with obstructions and varying object sizes. Utilizing real distance data from sensors to replace visually estimated data further enhances the realism of detections. We hope that our research findings highlight the potential of advanced algorithms and network architectures to improve the environmental perception and decision-making abilities of autonomous vehicles, leading to safer and more reliable smart transportation solutions.


\begin{thebibliography}{99}
\small

\bibitem{liu2020multi}
Liu, Y., Shi, L., Xu, W., Xiong, X., Sun, W., \& Qu, L. (2020, November). The multi-sensor fusion automatic driving test scene algorithm based on cloud platform. In \textit{2020 Chinese Automation Congress (CAC)} (pp. 5975-5980). IEEE.
\bibitem{liu2024kan}
Liu, Z., Wang, Y., Vaidya, S., Ruehle, F., Halverson, J., Soljačić, M., Hou, T. Y., \& Tegmark, M. (2024). Kan: Kolmogorov-Arnold networks. \textit{arXiv preprint arXiv:2404.19756}.
\bibitem{li2022bevdepth}
Li, Y., Ge, Z., Yu, G., Yang, J., Wang, Z., Shi, Y., Sun, J., \& Li, Z. (2022). BEVDepth: Acquisition of reliable depth for multi-view 3D object detection. arXiv preprint arXiv:2206.10092.
\bibitem{zhang2022beverse}
Zhang, Y., Zhu, Z., Zheng, W., et al. (2022). BEVerse: Unified Perception and Prediction in Birds-Eye-View for Vision-Centric Autonomous Driving. arXiv preprint arXiv:2205.09743.
\bibitem{huang2022bevdet4d}
Huang, J., \& Huang, G. (2022). BEVDet4D: Exploit Temporal Cues in Multi-Camera 3D Object Detection. \textit{arXiv preprint arXiv:2203.17054}.
\bibitem{tesla2022aiday}
Morando, A., Gershon, P., Mehler, B., \& Reimer, B. (2021). A model for naturalistic glance behavior around Tesla Autopilot disengagements. Accident Analysis \& Prevention, 161, 106348.
\bibitem{DepthNet2018}
Kumar, C.S., Bhandarkar, S.M., \& Prasad, M. (2018).
DepthNet: A recurrent neural network architecture for monocular depth prediction.
In \textit{Proceedings of the IEEE Conference on Computer Vision and Pattern Recognition Workshops} (pp. 283-291).
\bibitem{zhou2018voxelnet}
Zhou Y, Tuzel O. VoxelNet: End-to-end learning for point cloud based 3d object detection[C]//Proceedings of the IEEE conference on computer vision and pattern recognition. 2018: 4490-4499.
\bibitem{nabati2021}
Nabati, R., \& Qi, H. (2021). Centerfusion: Center-based radar and camera fusion for 3d object detection. In Proceedings of the IEEE/CVF Winter Conference on Applications of Computer Vision (pp. 1527-1536).
\bibitem{deng2021voxel}
Deng J, Shi S, Li P, et al. Voxel R-CNN: Towards high performance voxel-based 3d object detection[C]//Proceedings of the AAAI Conference on Artificial Intelligence. 2021, 35(2): 1201-1209.
\bibitem{lang2019pointpillars}
Lang A H, Vora S, Caesar H, et al. PointPillars: Fast encoders for object detection from point clouds[C]//Proceedings of the IEEE/CVF conference on computer vision and pattern recognition. 2019: 12697-12705.
\bibitem{Chen2017}
X. Chen, H. Ma, J. Wan, et al. ``Multi-view 3D object detection network for autonomous driving,'' in \textit{2017 IEEE Conference on Computer Vision and Pattern Recognition (CVPR)}, Honolulu, USA: IEEE, 2017, pp. 6526–6534.
\bibitem{Qian2021}
K. Qian, S. Zhu, X. Zhang, et al. ``Robust multimodal vehicle detection in foggy weather using complementary LiDAR and radar signals,'' in \textit{2021 IEEE/CVF Conference on Computer Vision and Pattern Recognition (CVPR)}, Nashville, USA: IEEE, 2021, pp. 444–453.
\bibitem{Singh2023} 
Singh, Apoorv. Vision-RADAR Fusion for Robotics BEV Detections: A Survey. arXiv, February 2023. doi:10.48550/arXiv.2302.06643.
\bibitem{huang2021bevdet}
Huang, J., Huang, G., Zhu, Z., \& Du, D. (2021). BEVDet: High-performance Multi-camera 3D Object Detection in Bird-Eye-View. arXiv preprint arXiv:2112.11790.
\bibitem{bansal2020pointillism}
Bansal, K., Rungta, K., Zhu, S., \& Bharadia, D. (2020). Pointillism: Accurate 3d bounding box estimation with multi-radars. Proceedings of the 18th Conference on Embedded Networked Sensor Systems.
\bibitem{Liu2022}
Z. Liu, H. Tang, A. Amini, et al. ``BEVFusion: Multi-task multi-sensor fusion with unified bird’s-eye view representation,'' arXiv:2205.13542, 2022.
\bibitem{Chadwick2019}
S. Chadwick, W. Maddern, P. Newman. ``Distant vehicle detection using radar and vision,'' in \textit{2019 International Conference on Robotics and Automation (ICRA)}, Montreal, Canada: IEEE, 2019, pp. 8311–8317.
\bibitem{decourt2022recurrent}
Decourt, C., Van Rullen, R., Salle, D., \& Oberlin, T. (2022). A recurrent cnn for online object detection on raw radar frames. ArXiv, abs/2212.11172.
\bibitem{li2022deepfusion}
Li, Y. W., Yu, A. W., Meng, T. J., et al. (2022). DeepFusion: Lidar-Camera Deep Fusion for Multi-Modal 3D Object Detection. In \textit{2022 IEEE/CVF Conference on Computer Vision and Pattern Recognition (CVPR)}, New Orleans, LA: IEEE, pp. 17161-17170.
\bibitem{liu2023petrv2}
Liu, Y., Yan, J., Jia, F., Li, S., Gao, A., Wang, T., \& Zhang, X. (2023). PETRv2: A Unified Framework for 3D Perception from Multi-Camera Images. In \textit{Proceedings of the IEEE/CVF International Conference on Computer Vision} (pp. 3262-3272).
\bibitem{li2024fastbev}
Y. Li, X. Zhang, H. Wang, J. Liu, and Z. Liu, "Fast-BEV: A Fast and Strong Bird's-Eye View Perception Baseline," in \textit{IEEE Transactions on Pattern Analysis and Machine Intelligence}, doi: 10.1109/TPAMI.2024.3414835.
\bibitem{li2022voxel}
Li, Y. W., Qi, X. J., Chen, Y. K., et al. (2022). Voxel Field Fusion for 3D Object Detection. In \textit{2022 IEEE/CVF Conference on Computer Vision and Pattern Recognition (CVPR)}, New Orleans, LA: IEEE, pp. 1110-1119.
\bibitem{wang2022sts}
Z. Wang, C. Min, Z. Ge, Y. Li, Z. Li, H. Yang, and D. Huang, “STS: Surround-view temporal stereo for multi-view 3D detection,” arXiv preprint arXiv:2208.10145, 2022.
\bibitem{qin2023supfusion}
Qin, Y., Wang, C. Q., Kang, Z. J., et al. (2023). SupFusion: Supervised Lidar-Camera Fusion for 3D Object Detection. In \textit{Proceedings of the IEEE/CVF International Conference on Computer Vision}, Paris, France: IEEE, pp. 22014-22024.

\end{thebibliography}
\end{document}